# Non-contact PPG Signal and Heart Rate Estimation with Multi-hierarchical Convolutional Network

**Bin Li [1], Panpan Zhang [1], Jinye Peng [1], Hong Fu [2,*]**

[1] School of Information Science and Technology, Northwest University, Xi'an, People's Republic of China.

[2] Department of Mathematics and Information Technology, The Education University of Hong Kong, Hong Kong.

lib@nwu.edu.cn(B.L.); zhangpanpan1k@163.com(P.Z.); pjy@nwu.edu.cn(J.P.); hfu@eduhk.hk

* Corresponding Author, E-mail: hfu@eduhk.hk

**Abstract:** Heartbeat rhythm and heart rate (HR) are important physiological parameters of the human body. This study presents an efficient multi-hierarchical spatio-temporal convolutional network that can quickly estimate remote physiological (rPPG) signal and HR from face video clips. First, the facial color distribution characteristics are extracted using a low-level face feature generation (LFFG) module. Then, the three-dimensional (3D) spatio-temporal stack convolution module (STSC) and multi-hierarchical feature fusion module (MHFF) are used to strengthen the spatio-temporal correlation of multi-channel features. In the MHFF, sparse optical flow is used to capture the tiny motion information of faces between frames and generate a self-adaptive region of interest (ROI) skin mask. Finally, the signal prediction module (SP) is used to extract the estimated rPPG signal. The heart rate estimation results show that the proposed network overperforms the state-of-the-art methods on three datasets, 1) UBFC-RPPG, 2) COHFACE, 3) our dataset, with the mean absolute error (MAE) of 2.15, 5.57, 1.75 beats per minute (bpm) respectively.

## 1. Introduction

Physiological signals caused by the periodic activity of the heart contain a variety of important vital signs, such as heart rate (HR), blood oxygen, and heart rate variability (HRV), which are considered to be important indicators of human health and considerably significant in various fields such as medical treatment [1], sports, and psychological testing [2]. Two common methods of measuring these indicators are electrocardiogram (ECG) and photoplethysmography (PPG). The ECG measures the electrical activations that lead to the contraction of the heart muscle, using electrodes attached to the body, usually at the chest, it is widely used in the medical field. PPG uses a small optical sensor combined with a light source to measure changes in light absorption by blood vessels caused by the cardiac cycle. This method is usually used for wearable monitoring devices such as sports bracelets [3]. However, both methods are based on contact measurement, which is inconvenient to users and unfriendly to people with extremely sensitive skin such as newborns or burn patients. Thus, many researchers have focused on developing non-contact measurement methods.

Non-contact physiological signal estimation approaches are mainly divided into two categories: (1) cardiography [4] based on periodic head movement, which aims to extract heartbeat information from the subtle head movement caused by the heart periodically injecting blood into the large blood vessels, and (2) remote photoplethysmography (rPPG), which aims to obtain heartbeat information by analyzing periodic changes in the skin's absorption of light caused by changes in blood volume and oxygen saturation in the blood vessels. The latter is more robust to changes in head movement and illumination. With the development of imaging technology, commercial cameras have met the video capture requirements for remote physiological signal measurement. Early research on HR measurement based on periodic skin changes [5] has achieved good results under normal ambient light conditions, which can be summarized as follows: (1) detection of regions of interest (ROIs) by face detection with a feature tracker, and calculation of raw signals by averaging the respective color channel of all pixels contained in the ROI of the frame; (2) acquisition of the plethysmographic signal using signal analysis methods such as filtering and dimensionality reduction; (3) estimation the average HR using frequency analysis and peak location. However, this change in light absorption by skin tissue due to blood volume is too weak to be observed by the eyes; it is also easily affected by head movement and light changes. Traditional feature extraction and filtering processes may lose important information and are unfriendly to data. In the era of big data, some data-driven methods such as deep learning have shown powerful modeling capabilities and have attracted increasing attention. Deep learning-based remote HR estimation has also achieved good results. However, all the aforementioned methods have one or the other of the following shortcomings: (1) design based on a two-dimensional (2D) spatial neural network, neglecting time information in video-based remote HR estimation; (2) not an end-to-end method, as the input is not the original facial video; (3) ROI detection is unstable when there is a large range of head movements in the video sequence; (4) longer measurement time, generally requiring 30s of facial video to calculate the average HR[6-8].

In this study, we propose an end-to-end three-dimension (3D) spatio-temporal convolutional network to efficiently recover the rPPG signal and average HR value from the original video clips. We add a multi-hierarchical feature fusion (MHFF) based attention module to make the network focus on rPPG-related features and reduce the impact of head movement and background noise on signal recovery.

The main contributions of our work include the followings:

1. An efficient end-to-end 3D spatio-temporal convolutional network with an MHFF-based attention model is proposed. The skin map label generated based on sparse optical flow effectively solves the influence of background noise and head movement.

2. Only 15 s face video clips are needed for efficient reconstruction of rPPG signal and accurate estimation of HR, and the experiments are conducted on three datasets to verify the effectiveness of the proposed network.

3. A new face video physiological parameters dataset FaceBio-v1 with annotated PPG and HR signal is presented, which contains 300 videos from 300 subjects.

The remainder of this paper is organized as follows. Section 2 reviews video-based remote physiological measurement algorithms in recent years. In Section 3, the proposed method is described in detail. The experimental results and discussion are presented in Section 4. Finally, Section 5 summarizes the study.

## 2. Related work

Previous methods for rPPG signal measurement and spatio-temporal network frameworks are briefly reviewed in the following two subsections.

### 2.1 Remote Photoplethysmograph Signal Measurement

The mainstream rPPG-based remote HR measurement methods can be divided into two types: the traditional method based on signal analysis and the data-driven method based on deep learning.

Verkruysse et al. [10] proposed an early rPPG-based remote physiological signal measurement method that requires the objects to remain stationary under ambient light. In these recordings, the ROIs were manually selected through the bounding box. From the pixels contained in the ROIs, the average value of the red-green-blue (RGB) color channels was calculated per frame as the original signal, and a band-pass filter was used to remove the noise. It was found that the green channel contains the strongest PPG signal; this was used in the initial work. To improve the manual bounding box, Poh et al. [5] used a face detector to track the object's face frame by frame, and used a 30s moving window and a bounding box containing the face as the ROI to achieve continuous measurement. Weighting and merging the color space produced better results than the green single-channel signal. CHROM [6] uses a standardized skin color configuration to perform white balancing operations on video frames and obtain a linear combination of chromaticity signals. Poh, McDuff, and Picard [3,10] introduced the blind source separation (BSS) method to estimate the rPPG signal as a linear combination of three channels, this improves the signal-to-noise ratio (SNR) and achieves the purpose of dimensionality reduction. They used independent component analysis (ICA) to identify the major components related to HR. In contrast to the ICA method applied to the entire ROI, Lam and Kuno [11] applied ICA at the patch level for HR estimation. In addition to ICA, principal component analysis (PCA) and constrain ICA (CICA) are also used to estimate the HR signal. Lan et al. [12] compared various linear and nonlinear techniques of BSS and found that Laplacian feature mapping could produce the best results. After the main components of the HR were obtained, fast Fourier transformation (FFT) was needed to obtain the spectrum diagram; the frequency with the highest response is estimated to be the corresponding HR frequency. To reduce the noise spectrum contained in the FFT, Kumar et al. [13] combined HR signals with different ROIs using frequency features as weights. All these methods require the assumption of a feasible frequency band for human HR. The common choice of the frequency band is [0.7 Hz, 4 Hz], which corresponds to an HR of 42 to 240 beats per minute (bpm) [14]. In addition, there are also some methods based on reflection models, such as converting RBG channels into other color spaces [7] to obtain more robust rPPG signals. However, these methods rely on prior knowledge of researchers, and their performance largely depends on the effect of manual data preprocessing and whether the environment meets the hypothesis conditions. Complex constraint conditions and weak generalization ability hinder the achievement of good results in a non-constrained environment with large noise such as ambient light change and head movement, by traditional methods; therefore, these methods have certain limitations in practical applications.

In the era of big data, data-driven learning-based methods have rapidly developed and provided better results in complex environments. Tulyakov et al. [15] divided the face into multiple ROI regions to obtain a time representation matrix and used matrix completion to purify HR-related signals. Wang et al. [16] proposed an object-related method that uses spatial subspace rotation to compute HR signals by providing a complete continuous sequence of the object's face. Hsu et al. [17] combined the frequency domain features of RGB channels and ICA components to obtain the representation of HR signals and used support vector regression to estimate HR. Based on this work, they proposed an improved method with a better effect [18]. After extracting rPPG signals using the traditional CHROM method [6], they used short-time Fourier transform to construct 2D time-frequency representation (TFR) sequences, and 2D TFR was used as the input of the VGG-16 model and the majority voting based network to achieve HR estimation through. To solve the problem of insufficient training data, Niu et al. [19] proposed a transfer learning method called Synrhythm, which first uses the synthesized HR signal to pre-train the model, and then transfers the pre-trained model to the real HR estimation task. As head movement had a higher effect on brightness than on chrominance, in a later study, Rhythmnet [20] converted the video sequence from an RGB channel to a YUV channel and computed a spatiotemporal map from ROIs as a representation of the HR signal. It then used a convolutional neural network (CNN)–recurrent neural networks (RNN) model incorporating a gating unit (GRU) to learn the mapping from the spatiotemporal map to HR. However, these methods still require a strict preprocessing stage and are based on preset ROIs for signal extraction while neglecting other facial areas. End-to-end deep learning methods have also been proposed. Spetlik, Cech, and Vojtěch [21] proposed an HR-CNN network that includes the extractor and HR estimator. The extractor is a 2D CNN with frequency constraints that extracts non-contact reflection PPG (NrPPG) signals from facial images. The HR estimator estimates the HR by calculating different loss functions of the NrPPG signals. Considering the influence of head movement on rPPG signal extraction, DeepPhys [7] extracted the relevant physiological signals from a skin reflex-based motion representation model and a CAN-based appearance model, according to the different frequency characteristics of different signals to extract the corresponding physiological signals from a 30s facial video. Tang et al. [22] performed skin pixel segmentation through CNN and extracted the HR signals of the detected skin areas through PCA and ICA signal analysis. Such skin-based signal extraction is conducive to continuous HR measurement in challenging environments, such as neonatal intensive care units (NICUs). Tsou et al. [23] proposed the Siamese-rPPG Network, which is a weight-sharing mechanism to extract rPPG signals from the cheeks and forehead respectively, it added the two branch signals to obtain face rPPG signals to avoid signal loss when one ROI is occluded. To effectively restrain noise features while maintaining the integrity of periodic physiological features, Niu et al [9] processed the video sequence into MSTmap as the network input and proposed a strategy of cross-verified feature disentangling (CVD) to extract physiological signals from interference signals through a 30 s facial video. In addition to data noise, such as environmental factors, the network structure is also one of the main factors affecting the prediction results. AutoHR [24] composed of a searched backbone and time difference convolution, achieved remote HR measurement by finding an appropriate network backbone through neural architecture search. Although 2D CNN achieved a good effect in spatial feature extraction, it neglects the interframe motion information of the time dimension. Some works first performed spatial pooling of RGB sequences and then projected them into a more representative color space [4,25]. Thus, temporal context-based normalization was needed to avoid the effects of noise such as movement and environmental changes. Wang et al. [26] proposed a 2D CNN-based two-stream approach for signal extraction and an LSTM network for signal refining. Meta-rPPG [27] proposes 2D CNN + BiLSTM spatio-temporal network for signal extraction, meta-learning approach for fast adaptation. The use of a 3D CNN-based network can integrate these steps and regard the time dimension as the third dimension. MTTS-CAN [28] proposes a multi-task time-shifted convolutional attention network with a dual-branch structure with a spatial attention module, which combines 2D-CAN and 3D-CAN to form Hybrid-CAN to maintain temporal modeling while efficiently utilizing 2D Convolution to extract spatial features. Yu et al. [29] proposed a 3D network for regressing rPPG signals from facial videos. ETA-rPPG [30] adds a temporal segmentation sub-network to solve the problem of redundant extraction of video information, and the joint attention mechanism strengthens feature extraction and eliminates noise. However, this method did not consider the influence of compressed video on signal recovery. Yu et al. [8] proposed a two-stage network containing STVEN for video enhancement and rPPGNet for rPPG signal recovery, and accurately recovered rPPG signals from a 30 s highly compressed facial video. In addition, Huang et al. [31] use 3D CNN + LSTM spatio-temporal network for signal extraction, besides 3D CNN, LSTM is used to further focus on temporal information. In addition to CNN, some researchers have also achieved good results in conjunction with GAN. Deep-HR [32] uses two GAN-style modules to enhance the detected ROI and remove noise, 2D CNN for signal extraction, and PulseGAN [33] uses a generative adversarial network model to post-process the estimated rPPG signal to reduce noise and improve output quality.

**2.2 Spatio-temporal Network Frameworks**

With the continuous development of machine vision, traditional 2D networks can no longer satisfy many video-based learning tasks (such as action recognition and video subtitles) because 2D CNN cannot capture the time context information of video media. Researchers have focused on the spatiotemporal network structure which can capture both temporal and spatial information and extract the local spatial features of the image while maintaining the time continuity of the video. At present, mainstream spatiotemporal network frameworks are divided into two categories. The first is a spatio-temporal network framework that combines a CNN and an RNN, for example, long-term short-term memory (LSTM) and convolutional LSTM, CNN feed-forward network is used for spatial modeling, and LSTM is used for temporal modeling. However, this structure causes difficulties in end-to-end optimization. The second category is the 3D CNN-based spatiotemporal network framework. Du et al. [34] first proposed a 3D convolutional network (C3D) that could simultaneously perform temporal and spatial modeling. In the subsequent development, various 3D CNN-based spatiotemporal network frameworks have been derived, which are widely used in video understanding and are more suitable for video-based remote physiological signal estimation depending on small facial skin tone changes [35].

In contrast to previous works, considering the comfort of the test subjects in practical applications. we further accelerated the signal reconstruction speed on the basis of ensuring the prediction accuracy. We rapidly and accurately reconstructed the physiological signals through the short-time facial video and calculated the 15s average HR. Feature maps at different levels focus on different information: low-level feature maps have rich spatial information, whereas high-level feature maps focus on semantic information. To consider both spatial and semantic information, we propose a multi-hierarchical convolutional network that integrates multi-level feature maps.

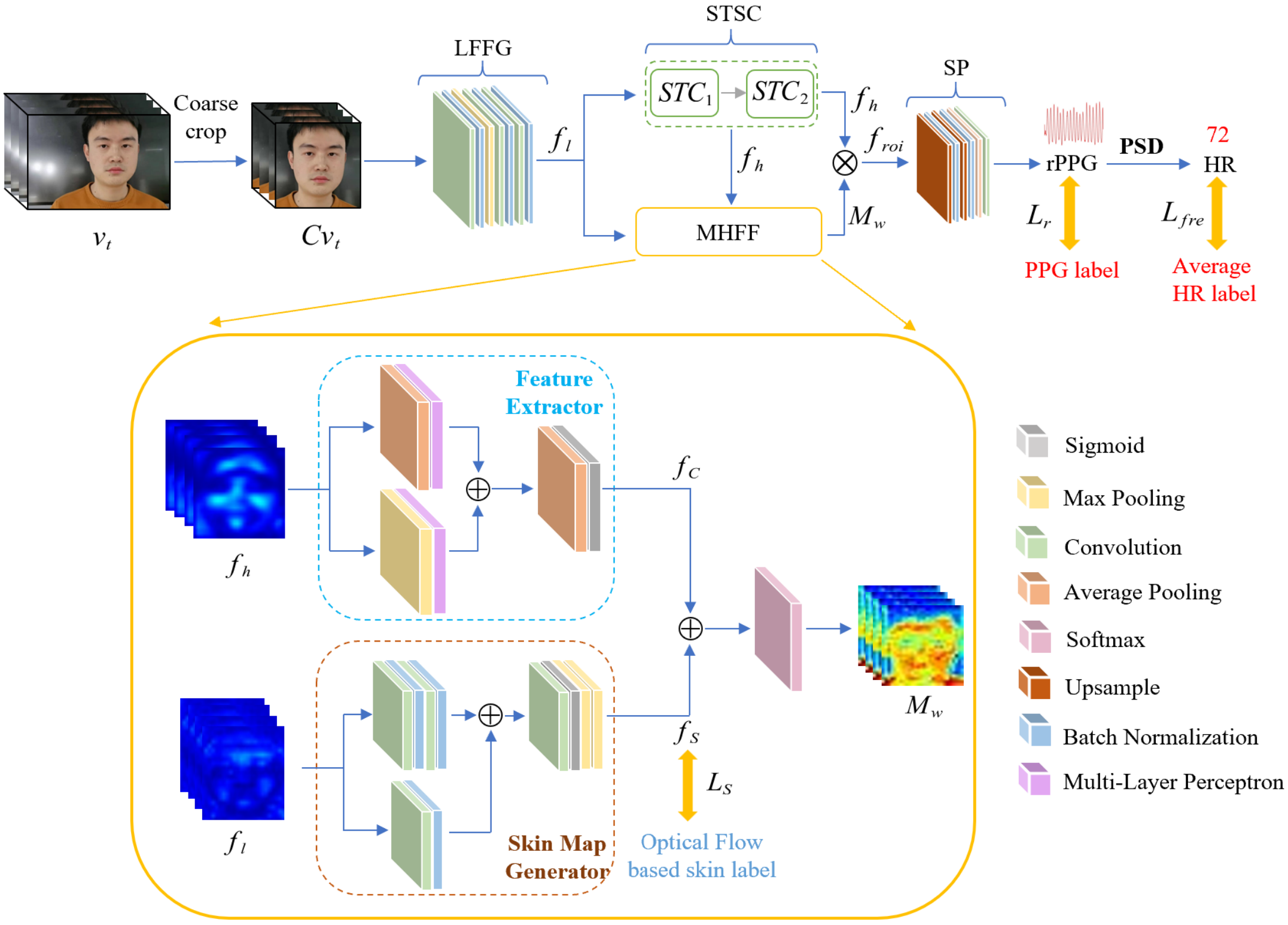

**Fig.1** The architecture of the proposed method. The top is the overall structure of the proposed network. The bottom is the structure of MHFF that include a C_feature extractor and a skin map generator.

## 3. Methodology

The proposed method aims to reconstruct rPPG from RGB facial videos by designing a 3D spatio-temporal convolutional network with multi-hierarchical fusion. As shown in Fig.1, the proposed network includes four modules: low-level face feature generation (LFFG), 3D spatio-temporal stack convolution (STSC), multi-hierarchical feature fusion (MHFF), and signal predictor (SP).

### 3.1. Data Preparation

As the variation range of facial skin's light absorption is very weak, some highly reflective backgrounds may affect the predicted results, so we used [36] cascading the ensemble of regression trees (ERT) algorithm [37] to approximately crop the redundant background of the video.

$$Cv_t = crop(v_t) \tag{1}$$

where $v_t$ is the original video frames, $Cv_t\left(t=1,2,...,T\right)$ is the coarse-cropped video frames. Then normalized video frames to $W \times H$ and $T$ frames are aggregated into a video stream as the network input $Cv_{t:t+T} \in \mathbf{R}^{C\times T\times W\times H}$, $C$, $T$, $W$, and $H$ is the channels ($C$=3 means red, green, and blue), the number of input frames, and the spatial sizes respectively.

### 3.2. Low-level Face Feature Generation

Low-level feature maps have rich spatial information that is beneficial for obtaining the facial color distribution in the image. Therefore, we propose a three-layer convolutional network model, LFFG, for superficial facial feature extraction. The input of LFFG is an RGB face video clip $Cv_{t:t+T} \in \mathbf{R}^{C\times T\times W\times H}$ with $T$ frames. The output of the LFFG is as follows:

$$f_l = conv_{2-3}\left(Max\left(conv_1(Cv_{t:t+T})\right)\right) \tag{2}$$

where $conv_1(\cdot)$ is the spatial convolution with a kernel size of $1\times5\times5$, $Max(\cdot)$ is the spatial max pool with $1\times2\times2$ kernel, $conv_{2-3}(\cdot)$ is the 3D spatio-temporal convolution with kernel size of $3\times3\times3$. After each convolution layer, there is a batch normalization and rectified linear unit (ReLU) activation layer. $f_l \in \mathbf{R}^{32\times T\times W\times H}$ are the low-level feature maps of facial video clips.

### 3.3. Spatio-temporal Stack Convolution

Temporal feature acquisition and multi-hierarchical spatiotemporal feature fusion are very important for rPPG signal recovery from facial video. Rich shallow spatial features obtained by LFFG are input into the STSC module for deeper feature extraction. STSC is composed of two serial convolutional blocks with similar structures, as shown in Fig.2.

$$f_{STC1} = STC_1(f_l),\ \ f_h = STC_2(f_{STC1}) \tag{3}$$

where $STC_i(\cdot)(i=1,2)$ is a convolutional block, and $f_{STC1}$ is the output of $STC_1$. Each convolution block consists of two 3D spatiotemporal convolution layers and a downsample layer. Kernel size of 3D spatio-temporal convolution layers are $3\times3\times3$. After each convolution layer, there is batch normalization and ReLU activation layer. To extract the temporal context information more effectively and reduce the redundancy and noise of the time dimension, the down-sample layer subsamples the time dimension and the space dimension through the kernel size of $2\times2\times2$ and 2 steps. The output of the STSC is a high-level feature map $f_h \in \mathbf{R}^{64\times\frac{T}{4}\times W\times H}$.

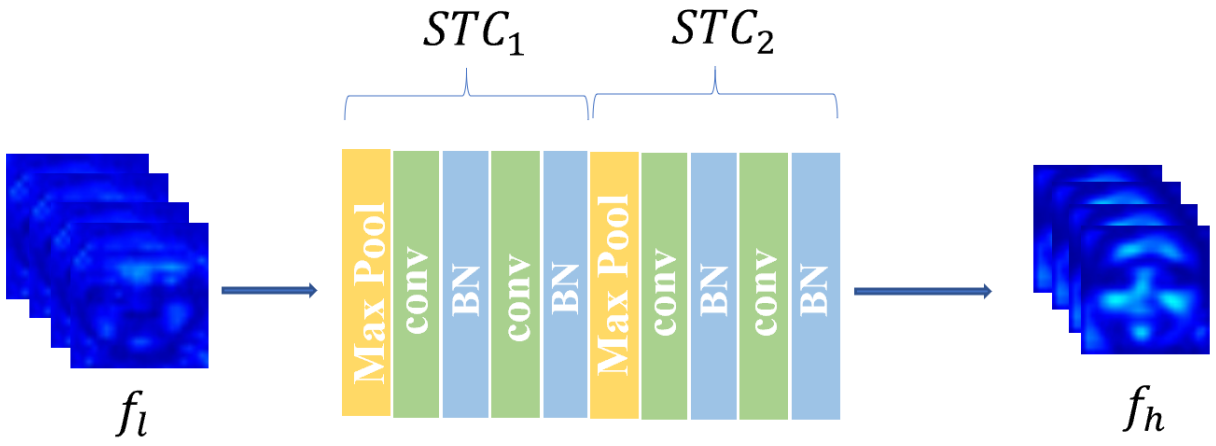

**Fig.2** The architecture of STSC. Includes two convolution blocks of similar structure.

To capture the inter-frame variation of the video clips, the output value feature map $f(p_t')$ at time *t*, position $p'$ is generated by stacking $2n+1$ consecutive feature maps $f(p_{t-n})$, $f(p_{t-n+1})$, ... , $f(p_{t+n-1})$, $f(p_{t+n})$ as shown in Equation (4) and Fig.3.

$$f(p_t') = \sum_{i=-n}^{n} f(p_{t+i}) \tag{4}$$

where $f(p_t')$ is the value of feature map at time *t*, position $p'$. Considering the computational efficiency and HR measurement accuracy, *n* is set to 1 in our experiments.

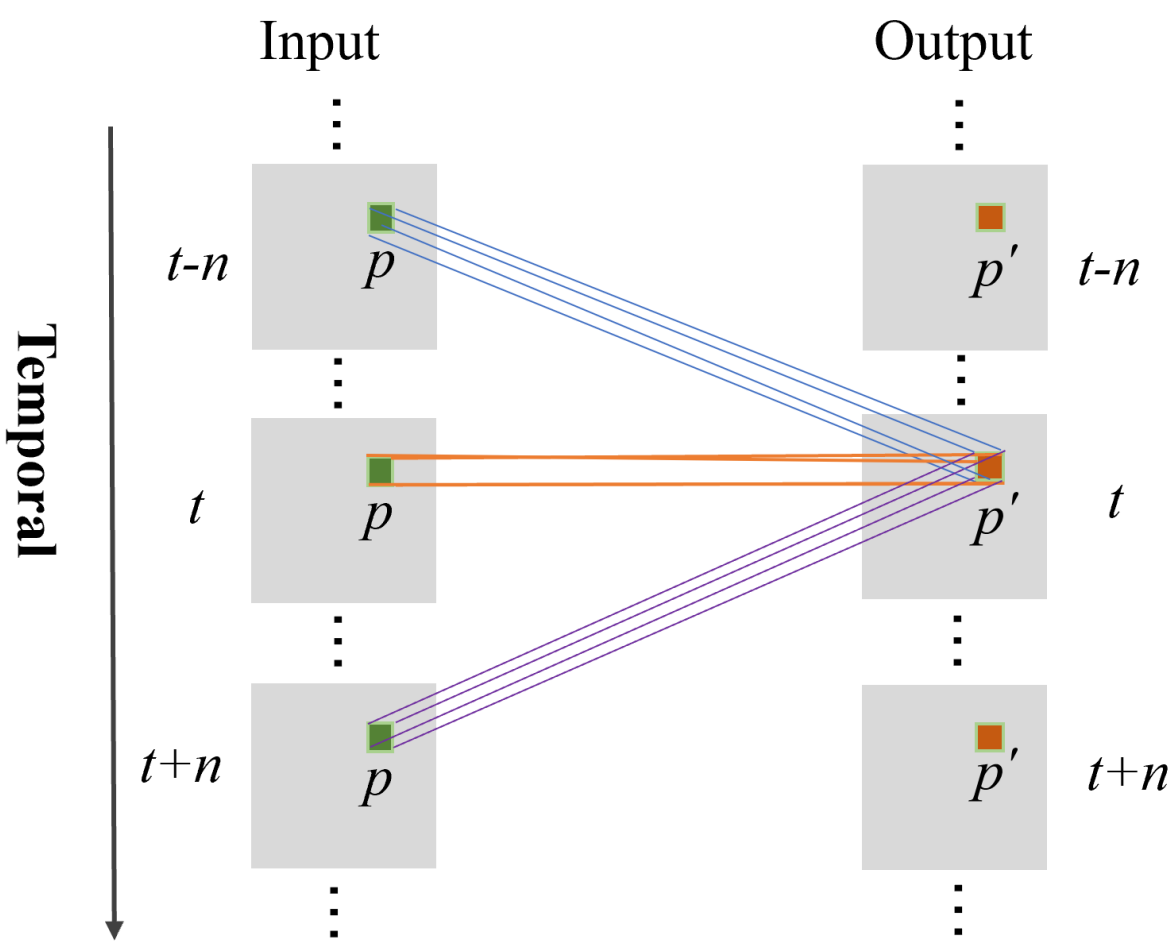

**Fig. 3** 3D spatio-temporal features stacking. The output value feature map $f(p_t')$ at time *t*, position $p'$ is generated by stacking $2n+1$ consecutive feature maps $f(p_{t-n})$, $f(p_{t-n+1})$, … , $f(p_{t+n-1})$, $f(p_{t+n})$.

### 3.4. Multi-hierarchical Feature Fusion

To retain more effective information and make the network focus on feature extraction related to the rPPG signal, we propose the MHFF structure to generate and fuse multi-hierarchical features, as shown in Fig.1. MHFF includes a parallel C_feature extractor and a skin map generator based on the residual structure.

**C_feature extractor:** Because the original R, G, and B channels carry different physiological signals. After multiple convolutions, the number of channel feature maps has been expanded from 3 to 64. To find more stable channel feature maps for rPPG signal reconstruction and HR estimation, we send the high-level feature map $f_h$ generated by STSC to the C_feature extractor for channel-wise feature extraction. The purpose of the C_feature extractor is to find the optimal weight combination mode among the channels feature map.

$$f_C = \partial\left(\xi\left(Max(f_h)\right)+\xi\left(Avg(f_h)\right)\right) \tag{5}$$

where $\xi(\cdot)$ includes two 2D convolution layers with a $1 \times 1$ kernel and a ReLU activation layer, $Max(\cdot)$ is the max pool with an output size of 1, $Avg(\cdot)$ is the average pool with an output size of 1, $\partial(\cdot)$ includes a spatial average pool and a sigmoid classifier. The output of the C_feature extractor is a channel-wise feature map $f_C \in \mathbb{R}^{\frac{T}{4} \times W \times H}$.

**Skin Map Generator:** To further emphasize the spatial ROI of the feature map, a skin map generator is proposed to perform spatial-wise feature extraction on the low-level feature map $f_l$ that includes rich spatial information to obtain the skin map $f_S$:

$$f_S = \sigma\left(conv(\phi(f_l))\right) \tag{6}$$

where $\phi(\cdot)$ is a convolutional block with a residual structure, which includes two backbone convolution layers with $1\times3\times3$ kernels and a residual convolution with a $1\times1\times1$ kernel. $conv(\cdot)$ is a spatial convolution with a kernel of $1\times3\times3$, $\sigma(\cdot)$ includes a sigmoid classifier and two down-sample layers of $2\times2\times2$ kernels with 2 steps. The output of the skin map generator is a skin map $f_S \in \mathbb{R}^{\frac{T}{4} \times W \times H}$.

Then, the C_feature extractor generates a weight mask by feature fusion of the low-level skin map $f_S$ and high-level channel-wise feature map $f_C$:

$$M_w = \psi(f_C + f_S) \tag{7}$$

where $\psi(\cdot)$ is a softmax classifier and the weight mask is $M_w \in \mathbb{R}^{\frac{T}{4} \times W \times H}$.

Finally, the high-level feature map $f_h$ is multiplied by the weight mask $M_w$ by channels.

$$f_{roi}^i = f_h^i \cdot M_w \tag{8}$$

where $f_h^i \in \mathbf{R}^{\frac{T}{4} \times W \times H}$ $(i = 1, 2, ..., 64)$ is the high-level feature map of the *i*-th channel, $f_{roi}^i$ is the hybrid feature map of *i*-th channel, and the final output of the MHFF is $f_{roi} \in \mathbb{R}^{64 \times \frac{T}{4} \times W \times H}$.

**3.5. Signal Predictor**

To reduce the redundancy of the time dimension, we inserted two subsamples in the feature extraction stage. Before feature aggregation, two consecutive up-samplings of the time dimension on the feature map are performed to restore the original time length.

$$f_{roi}^{up} = \rho(f_{roi}) \tag{9}$$

where $\rho(\bullet)$ is a transposed convolution with a kernel size of $4 \times 1 \times 1$. The stride is set as $2 \times 1 \times 1$ and the result of up-sampling is $f_{roi}^{up} \in \mathbb{R}^{64 \times T \times W \times H}$.

After multiple convolution operations, a multichannel spatio-temporal feature representation stream is formed. Spatial global average pooling is performed on $f_{roi}^{up}$ to aggregate the features of the spatial dimension and introduce an independent channel-wise convolution filter to project the multichannel spatio-temporal representation stream into the rPPG signal space.

$$rPPG = conv_C\left(Avg_G\left(f_{roi}^{up}\right)\right) \tag{10}$$

where $conv_C(\cdot)$ is the channel-wise convolution with a $1 \times 1 \times 1$ kernel, $Avg_G(\cdot)$ is a spatial global average pooling with an output size of $T \times 1 \times 1$. The output of the SP is an rPPG signal.

The rPPG signal contains a variety of physiological information. HR is a powerful predictor of the most common major medical events and is widely used in clinical medicine and daily health monitoring. The rPPG-based HR estimation is mainly divided into time-domain and frequency-domain methods. In the case of a short video sequence, we choose the frequency-domain method with higher accuracy. The frequency spectrum corresponding to the HR can be obtained through a frequency analysis of the rPPG signal. The HR can be obtained using the following formula.

$$HR = 60 \cdot \gamma_{hr} \tag{11}$$

Where $\gamma_{hr}$ is the frequency corresponding to the index with the highest spectral power.

### 3.6. Loss Function

To accurately recover the rPPG signal curve, we calculated both the time domain and frequency domain losses. Because both the ground truth PPG signal (from devices on fingers) and the predicted rPPG signal (from facial video) measure the blood volume changes using the optical method, the PPG and rPPG signals are similar. Because the ranges of PPG and rPPG signal values are not the same, compared to commonly used point-wise intensity errors such as mean square error, the Pearson product-moment correlation coefficient (Pearson correlation) can better guide the network to fit the curve tendency to accurately locate the peak position. The Pearson correlation ranges between –1 and 1. When the linear relationship between two variables is positively correlated, the correlation coefficient approaches 1, whereas it tends to –1 when the linear relationship between two variables is negatively correlated. If the correlation coefficient is equal to 0, there is no linear correlation between the variables. To make the predicted rPPG signal curve more similar to the ground truth, we used a negative Pearson correlation to calculate the loss in the time domain:

$$L_r = 1 - \frac{T\sum_{i=1}^{T} x_i y_i - \sum_{i=1}^{T} x_i \sum_{i=1}^{T} y_i}{\sqrt{\sum_{i=1}^{T} x_i^2 - \left(\sum_{i=1}^{T} x_i\right)^2}\sqrt{T\sum_{i=1}^{T} y_i^2 - \left(\sum_{i=1}^{T} y_i\right)^2}} \quad (12)$$

where T is the length of video clips, $x_i$ is the *i*-th predicted rPPG signal, $y_i$ is the *i*-th ground truth PPG signal.

Inspired by SNR loss, we treated HR estimation as a classification task through frequency transformation and introduced cross-entropy loss to calculate the loss in the frequency domain:

$$L_f = CE\left(PSD(X), HR_{gt}\right) \quad (13)$$

where $PSD(X)$ represents the power spectral density of the predicted rPPG signal, $HR_{gt}$ is the average HR value of the ground truth, and $L_f = CE(X, Y)$ is used to calculate the cross-entropy loss between the predicted value and the ground truth.

To avoid the influence of non-face areas, we also introduced the binary cross-entropy loss of the skin map $f_S$ and binary skin label $S_l$:

$$L_S = BCE(f_S, S_l) \tag{14}$$

where $BCE(X, Y)$ calculates the binary cross-entropy loss of the predicted value $X$ and the ground truth $Y$.

When obtaining binary skin labels, an optical flow field is introduced to generate a stable ROI. To accurately locate the ROIs of the human face while reducing time complexity, 30 major ROI landmarks in $Cv_0$ (the first frame) are selected to roughly divide the face region into five parts: forehead, eyes, nose, mouth, and cheeks. To self-adapt to changes in ROIs position between frames, the pyramid LK algorithm [38] is selected to track the 30 landmarks. If no face in the current frame, the previous frame is used to replace the current frame. We construct the L-layer face image pyramid as shown in Fig. (4).

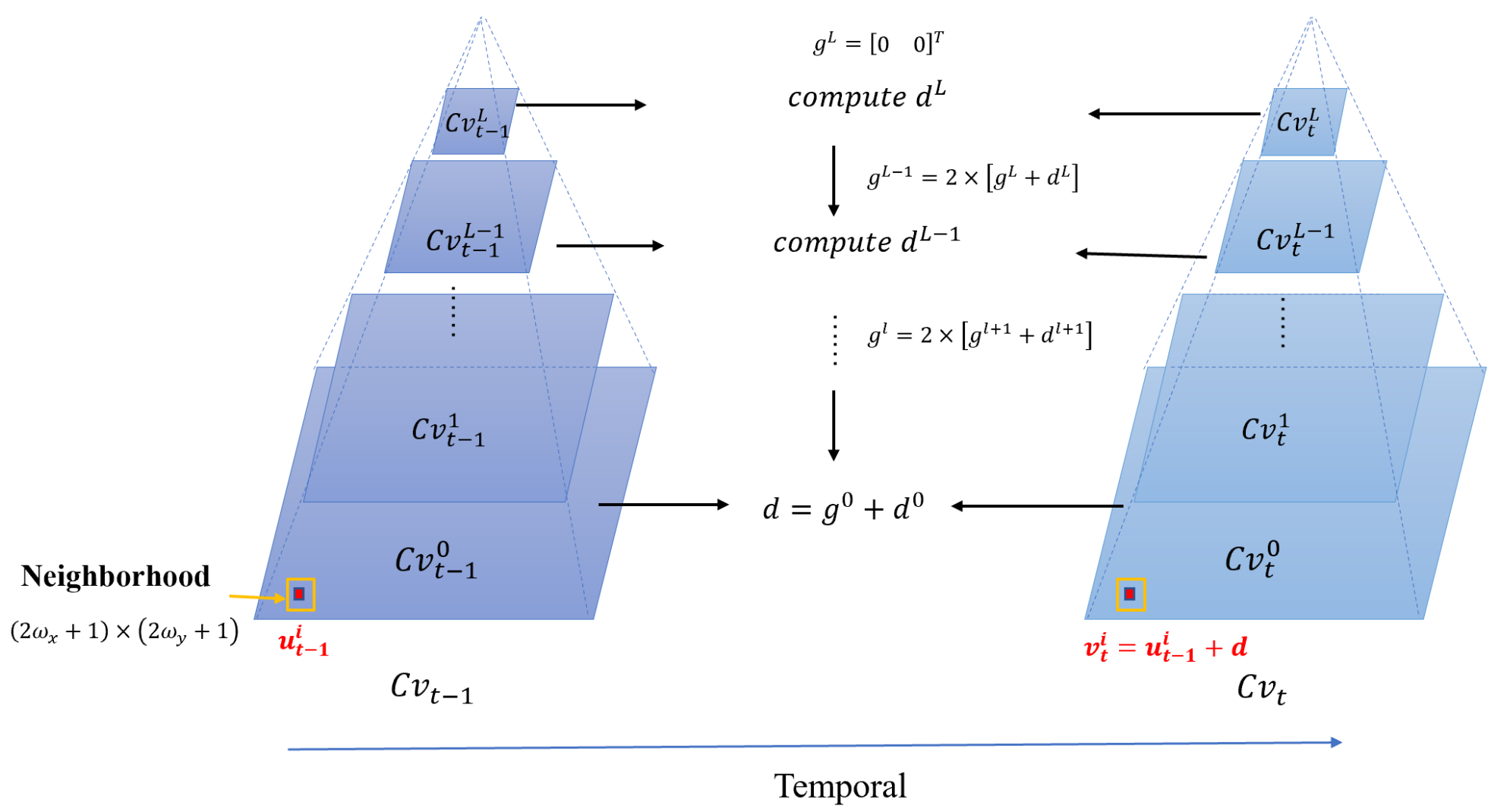

Fig.4. Pyramid Facial Frames.

In Fig. (4), $Cv_{t-1}^l$, $Cv_t^l$, and $l = (0,1,\ldots,L)$ represent the previous and current frame at $l$-th pyramid layer respectively. When $l = 0$, $Cv_t^0$ stands for the original frame $t$, and the size of the next layer is reduced to the half of the current layer. The gray value of the position $X = [x \quad y]^T$ in the video frame is denoted as $Cv(X) = Cv(x, y)$. The *i*-th facial landmark position in $Cv_{t-1}$ is marked as $u_{t-1}^{i} = \left[u_x^i \quad u_y^i\right]^T (i = 0,\ldots,29)$. The purpose of tracking is to locate the relevant landmark point $v_t^i = u_{t-1}^i + d = \left[u_x^i + d_x \quad u_y^i + d_y\right]^T$ in the next frame $Cv_t$ and minimize the error between $Cv_{t-1}\left(u_{t-1}^i\right)$ and $Cv_t\left(v_t^i\right)$, where the vector $d = [d_x \quad d_y]^T$ is called the optical flow of point *u* and *v*. From the top-layer *L* to the bottom layer 0 of the pyramid, the optimal value of optical flow is obtained by minimizing the grayscale matching error. Due to the spatial consistency assumption of the LK optical flow method, a neighborhood of size $(2\omega_x + 1) \times \left(2\omega_y + 1\right)$ ($\omega_x$ and $\omega_y$ are integers), centered on *u*, is taken as the integration window. $g^l = \left[g_x^l \quad g_y^l\right]^T$ is the optical flow of the *l*-th layer, and the optical flow of top layer *L*-th is initialized to $g^L = [0 \quad 0]^T$. According to [38], the residual optical flow $d^L$ of the *L*-th pyramid layer is calculated using the Taylor series expansion of formula (15).

$$d^L = min\left(\sum_{x=u_x^L-\omega_x}^{u_x^L+\omega_x} \sum_{y=u_y^L-\omega_y}^{u_y^L+\omega_y} \left(Cv_{t-1}^L(x,y) - Cv_t^L\left(x+g_x^L+d_x^L, y+g_y^L+d_y^L\right)\right)^2\right) \tag{15}$$

The initialization optical flow $g^{L-1} = 2 \times [g^L + d^L]$ is generated to obtain the residual optical flow of the (*L*-1)-th layer $d^{L-1} = \left[d_x^{L-1} \quad d_y^{L-1}\right]^T$. The above operations are iterated from *L*-th to 0-th to obtain the initial optical flow $g^0$ and the residual optical flow $d^0$ of the bottom layer $Cv_{t-1}^0$. Therefore, the optical flow of the frame *t* is $d = g^0 + d^0$, and the projection of the previous frame $Cv_{t-1}$ coordinate point $u_{t-1}^i$ to the current frame $Cv_t$ is $v_t^i = u_{t-1}^i + d$.

The face landmarks in the video frame and the corresponding binary skin labels are shown in Figure 5. The landmarks selected by the blue box in Fig. 5(a) are the 30 valid tracking points. There is a relative movement between the target and the background when the face is offset. The effective face ROI can be updated through sparse optical flow, which also reduces background noise and captures tiny inter-frame motion information. The skin labels are obtained to assist the skin map generator to generate a high-quality skin map as shown in Fig. 5(b), and the algorithm flow is listed in Table 1.

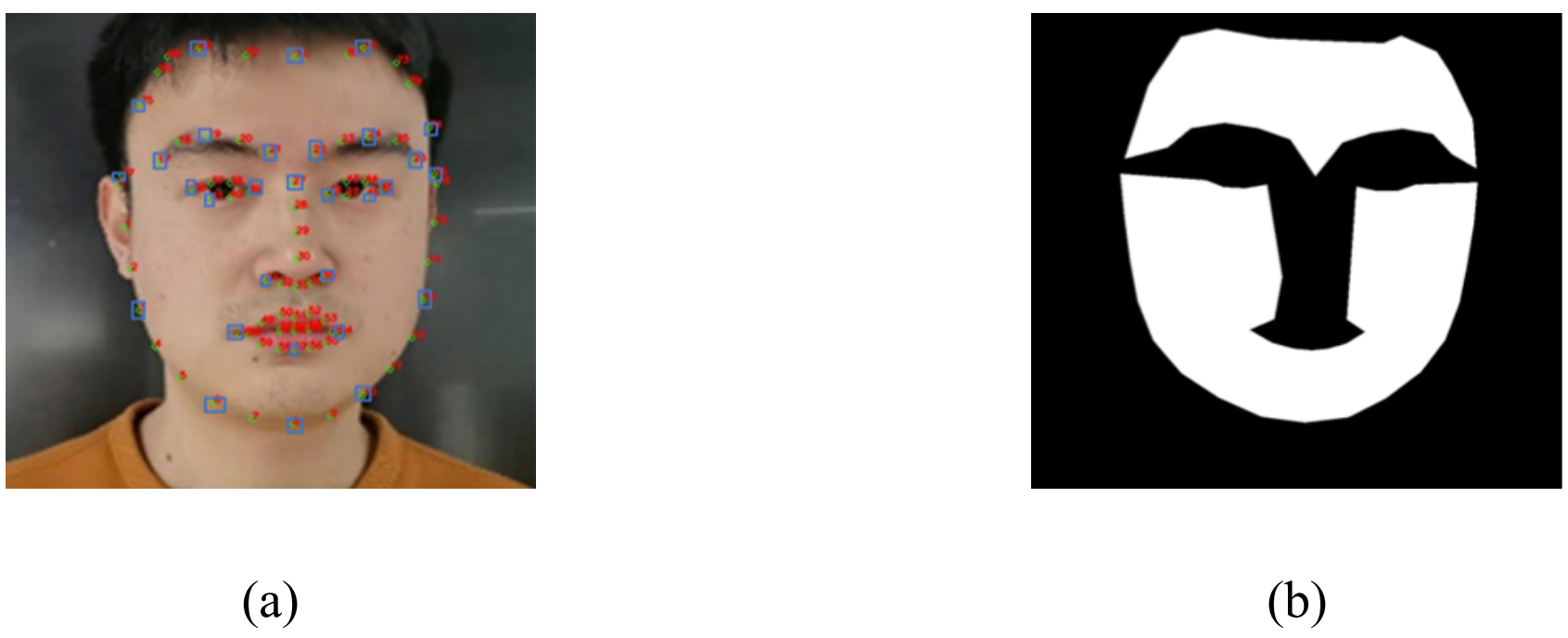

(a) (b)

**Fig.5** (a) Tracked facial landmarks. (b) Binary skin label

**Table. 1** Framework of Optical Flow Based Skin Label

| **Algorithm 1** Framework of Optical Flow Based Skin Label |
| --- |
| Input: RGB video frames $Cv_t \in [Cv_0,\cdots,Cv_T]$ |
| Output: skin label $S_l^t \in [S_l^0,\cdots,S_l^T]$ |
| 1: 30 facial landmarks $p_0 = \{p_0^0,\ldots,p_0^i,\ldots,p_0^{29}\}$ in $Cv_0$ are selected as the initial tracking points |
| 2: for $t = 1$ to $T$ do |
| 3: for $i = 0$ to 29 do |
| 4: $p_i^t = p_{t-1}^i + d$ // With sparse optical flow, the optical flow $d$ and coordinate point $p_i^t$ are obtained |
| 5: end for |
| 6： The ROIs of the forehead and cheeks are generated by facial landmarks |
| 7: for $(x, y)$ in $Cv_t$ do: |
| 8: if $(x, y)$ in ROIs: |

9: $S_l^t(x,y) = 255$

10: else $S_l^t(x,y) = 0$ // Binarization

11: end for

12: end for

13: return $S_l^t \in [S_l^0,\cdots,S_l^T]$

The final overall loss is calculated as:

$$Loss = \alpha L_r + L_f + \beta L_S \tag{16}$$

where $\alpha$ and $\beta$ are the weight coefficients to balance the loss.

## 4. Experiment results

### 4.1 Datasets

**Our dataset** (FaceBio-v1) contained a total of 300 VIS videos with a frame rate of 30 fps from 300 objects at the age of 18-26 years. The length of each video was 1 min with a pixel resolution of 1920×1080. These videos were collected by an Honor v30 mobile phone in a well-lit environment. Before the recording, the subject sat 60cm in front of the camera and remained resting for 5 minutes. The brightness of the facial region was kept between 200 and 400 lumens by a circular filling lamp. During the recording, subjects were allowed to perform natural facial expressions and head movements, but no talking was permitted. Physiological signals were collected by BIOPAC MP160, including the average HR, respiratory rate, $SpO^2$, ECG signal, and blood volume pulse (BVP) wave of each subject. The physiological signal sampling rate was 1000 Hz.

**The UBFC-RPPG** database contained 42 videos from 42 subjects [15]. The videos were recorded using a simple low-cost webcam (Logitech C920 HD Pro) at 30 fps with a resolution of 640×480 pixels in an uncompressed 8-bit RGB format. A CMS5OE transmissive pulse oximeter was used to obtain the ground truth PPG waveform and HR. During the recording, the subject sat in front of the camera (approximately 1 m away from the camera) with his/her face visible. All experiments were conducted indoors with varying amounts of sunlight and indoor illumination.

**The COHFACE** dataset contained 160 videos with high compression rates from 40 subjects (12 women and 28 men) [39]. Each of the subjects contributed four one-minute videos: two videos in well-lit conditions, and the other two captured under natural light. The videos were recorded using a Logitech HD C525 with a resolution of 640×480 pixels and a frame rate of 20 fps. Each subject wore a contact PPG sensor to obtain the BVP data.

### 4.2 Training settings

The facial videos and corresponding physiological signals were synchronized before the training. In our dataset, we set a variety of video sequence lengths $T = \{150, 300, 450, 600\}$ to obtain the best results, and the corresponding video durations were 5 s, 10 s, 15 s, and 20 s. The spatial dimensions of the coarse-cropped video frames were all standardized as $112 \times 112$ pixels. Our proposed method was implemented using the Pytorch framework and performed on an Intel(R) Core(TM) i7-9700 computer with 32 GB RAM and an NVIDIA GeForce RTX 2080Ti GPU. Adam optimizer with an initial learning rate of 0.0001 was used for training. The maximum epoch number for training was set to 80 for the experiments on our dataset.

### 4.3 Evaluation Metrics

We use the following four evaluation metrics to measure the performance of the proposed method:

(1) **Pearson correlation coefficient (r)**: The Pearson correlation coefficient varies between −1 and 1. Changes in the position and scale of the two variables do not cause changes in the coefficient. When the correlation coefficient approaches 1, it indicates that the two variables have a strong positive correlation, which can be used to measure the correlation between the predicted HR and ground truth.

$$r(x,y) = \frac{\sum_{i=1}^{n}(x_i-\bar{x})(y_i-\bar{y})}{\sqrt{\sum_{i=1}^{n}(x_i-\bar{x})^2}\sqrt{\sum_{i=1}^{n}(y_i-\bar{y})^2}} \quad (16)$$

(2) **Mean absolute error (MAE):** MAE is a linear score in which all individual differences have the same weight on the average value. It is used to measure the average value of the absolute error between the predicted HR and HR ground truth.

$$MAE = \frac{\sum_{i=1}^{n}|x_i-y_i|}{n} \quad (17)$$

(3) **Root mean square error (RMSE)**: The RMSE represents the sample standard deviation of the difference between the predicted value and the ground truth (called residual). The penalty for high errors is higher.

$$RMSE = \sqrt{\frac{\sum_{i=1}^{n}(x_i-y_i)^2}{n}} \quad (18)$$

(4) **Error standard deviation ($SD_e$)**: $SD_e$ is used to measure the dispersion degree of the absolute error, which can more intuitively reflect the stability of the method.

$$SD_e = \sqrt{\frac{\sum_{i=1}^{n}(err_i-\overline{err})^2}{n}} \quad (19)$$

Among the aforementioned evaluation metrics, $x_i$ represents the predicted HR, $\overline{x}$ represents the average predicted HR of all predicted HRs, $y_i$ represents the HR ground truth, $\overline{y}$ represents the average HR ground truth of all ground truths, $err$ represents the absolute error between the predicted HR and ground truth, and $\overline{err}$ represents the average absolute error.

**4.4 Intra-dataset Testing**

We first performed intra-dataset testing on our dataset and compared the accuracy of HR estimation with the state-of-the-art methods, as shown in Table 2. To ensure comparison fairness, we randomly set 240 subjects for training, 60 subjects for testing on our dataset, and the input video length is 15s for all learning-based methods.

Table2. The HR estimation results by the proposed method and several state-of-the-art methods on our dataset.

| Method | MAE↓ | RMSE↓ | $SD_e$↓ | r↑ |
|---|---|---|---|---|
| Green [10] | 6.59 | 11.91 | 9.92 | 0.53 |
| ICA [5] | 6.19 | 9.86 | 7.67 | 0.45 |
| POS [25] | 5.95 | 10.56 | 8.72 | 0.54 |
| CHROM [6] | 4.15 | 8.03 | 6.88 | 0.65 |
| PhysNet [29] | 2.41 | 4.39 | 3.91 | 0.88 |
| HR-CNN [21] | 3.47 | 5.46 | 4.28 | 0.73 |
| MSTmap+CVD [9] | 2.59 | 7.40 | 2.98 | 0.76 |
| Proposed Method | **1.75** | **3.43** | **2.95** | **0.93** |

It can be seen from the results that our proposed method achieved the expected results with an MAE of 1.75 bpm, RMSE of 3.43 bpm, SD of 2.95 bpm, and r of 0.93. The results of all four indicators were better than the results of traditional and deep-learning-based state-of-the-art methods. These results also indicate that on larger datasets, deep learning-based methods (PhysNet, proposed method) can achieve better performance. By reducing the MAE, the error distribution was also more concentrated, which is reflected in the smaller $SD_e$. As shown in Fig.6, the peaks and troughs of the rPPG signal we predicted are well aligned with the ground truth PPG signal, which meets the prediction requirements of multiple physiological signals such as respiratory rate and HRV with considerable application potential. To intuitively reflect the impact of facial video length on the HR prediction of our proposed method, we fix the model parameters and Fig.7(a) shows how different video lengths affect the HR estimate error (RMSE) on our dataset. The RMSE gradually decreases as the video length increases. However, we found that the RMSE increased when the video length increased to 20s. This is due to the large feature redundancy of facial ROIs between the adjacent video frames. While increasing the video length can help the network better acquire the spatiotemporal features of facial color tiny changes, the redundant information causes the network to be easily overfitted and lose focus on facial ROIs during training, resulting in a lower HR estimation accuracy. The feature maps output by 13s, 15s, 18s, and 20s videos are shown in Fig.7(b), It can be seen that when the video length reaches 20s, the feature extraction ability decreases.

The error distribution of the proposed method for different input video lengths is presented in Fig.8. The scatter plot of the predicted HR and the ground truth is shown in Fig.8(a), it can be seen that these parameters are highly positively correlated. As shown in Fig.8(b), most samples (88.5%) had an absolute error of less than 3 bpm; 92.8% of samples had an absolute error of less than 5 bpm, indicating that our proposed method has strong stability.

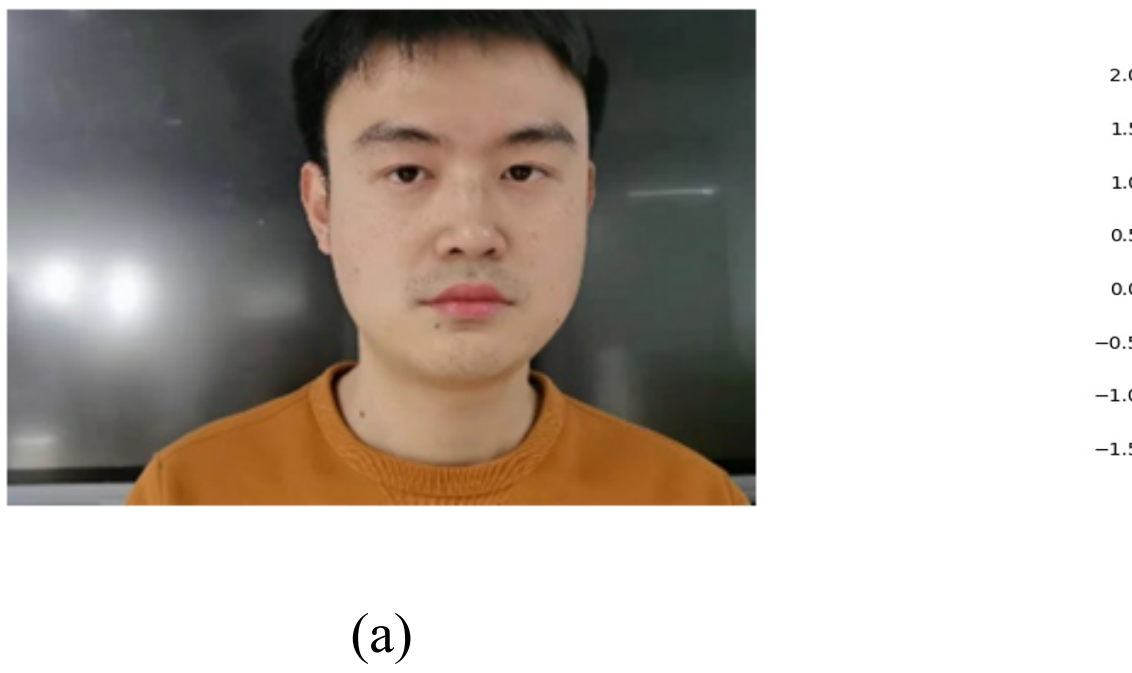

(a)

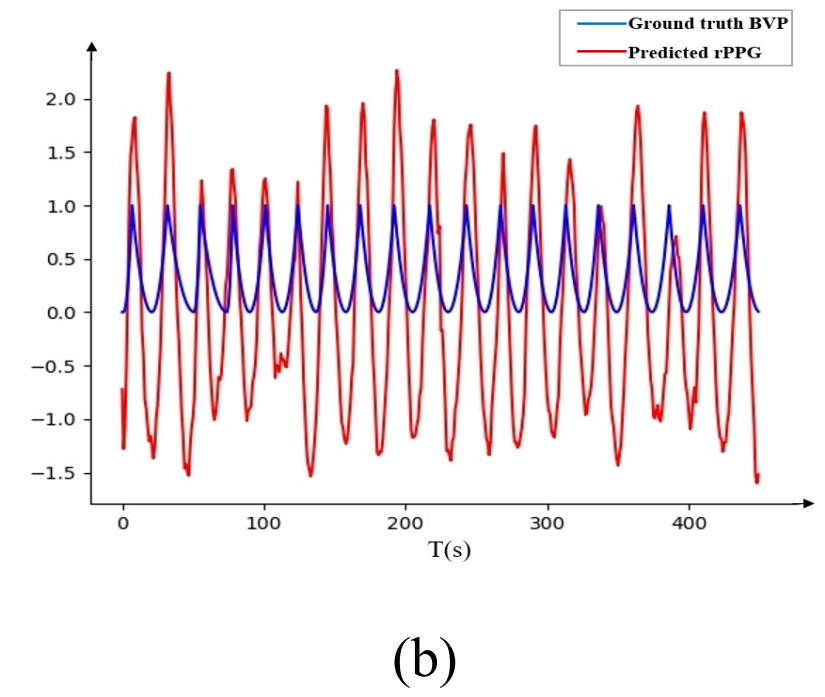

(b)

**Fig.6** An example from our dataset. (a) The original video frame in our dataset. (b)Example waves of the estimated and the ground truth signal, the blue wave is the ground truth PPG signal, and the red wave is the predicted rPPG signal.

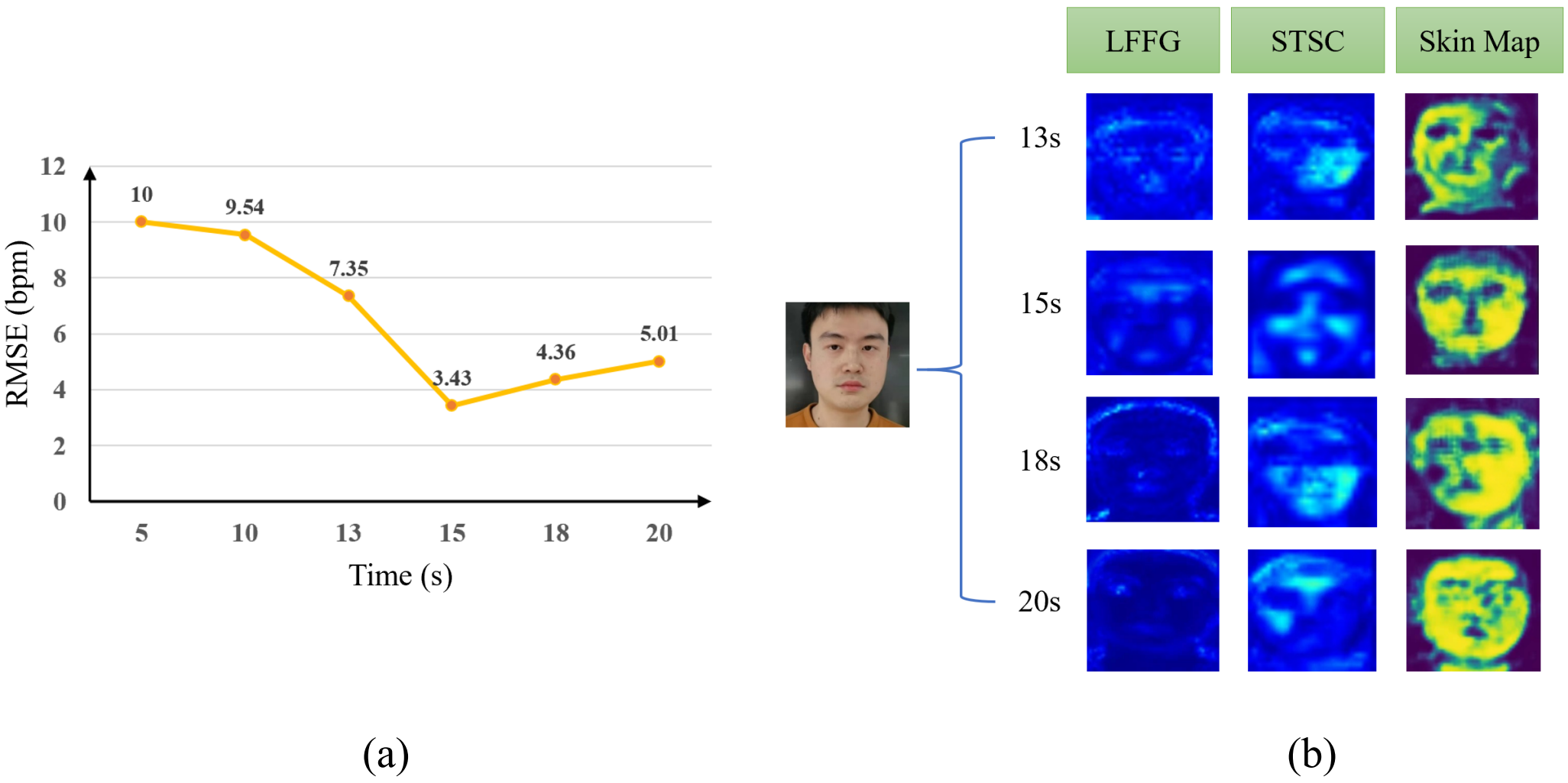

**Fig.7** (a) Evaluation of the HR prediction with different video lengths. (b) The feature maps are generated with fixed model parameters from 13s, 15s, 18s, and 20s videos.

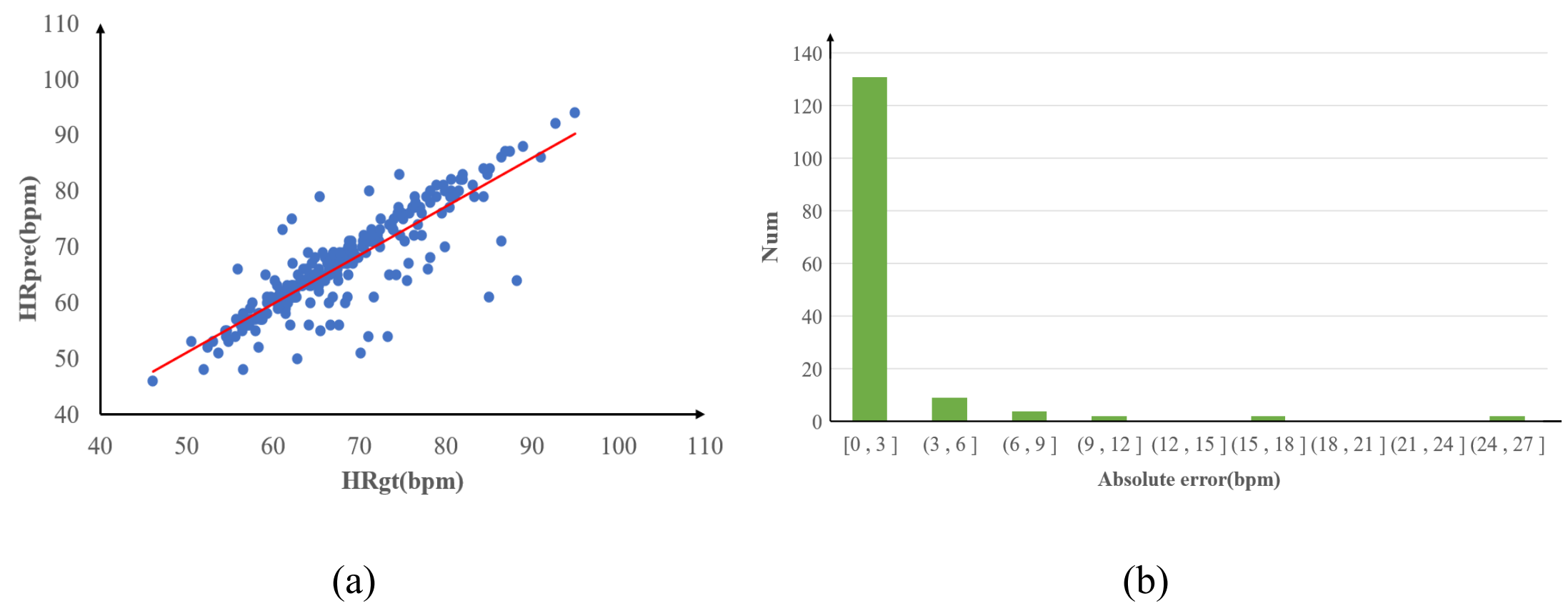

**Fig.8** (a) The scatter plot of the ground truth and the predicted HR. (b) The estimation error distribution of our proposed method.

### 4.5 Cross-dataset Testing

To evaluate the generalization ability of the proposed method, the model trained on our dataset is directly tested on the UBFC-RPPG and COHFACE datasets, and the length of input video clips is 15s. The test results of the proposed method and state-of-the-art methods are presented in Tables 3 and 4.

Table 3. The HR estimation results by the proposed method and several state-of-the-art methods on the UBFC-RPPG dataset

| Method | MAE↓ | RMSE↓ | $SD_e$↓ | r↑ |
| --- | --- | --- | --- | --- |
| Green | 10.20 | 20.60 | 20.20 | 0.56 |
| ICA | 8.43 | 18.80 | 18.60 | 0.68 |
| POS | 4.12 | 10.50 | 10.40 | 0.87 |
| CHROM | 10.60 | 20.30 | 19.10 | 0.59 |
| PhysNet | 3.63 | 5.29 | 3.85 | 0.94 |
| HR-CNN | 4.31 | 5.08 | 4.67 | 0.85 |
| MSTmap+CVD | 7.85 | 9.05 | 11.43 | 0.79 |
| Proposed Method | **2.15** | **3.82** | **3.15** | **0.97** |

Table 4. The HR estimation results by the proposed method and several state-of-the-art methods on the COHFACE dataset

| Method | MAE↓ | RMSE↓ | $SD_e$↓ | r↑ |
| --- | --- | --- | --- | --- |
| Green | 14.57 | 17.26 | 15.67 | 0.12 |
| ICA | 12.24 | 15.67 | 14.28 | 0.24 |
| POS | 13.43 | 17.05 | 12.89 | 0.07 |
| CHROM | 7.80 | 12.45 | 8.23 | 0.26 |
| PhysNet | 8.82 | 11.85 | 7.90 | 0.47 |
| HR-CNN | 8.10 | 10.78 | 8.87 | 0.29 |
| MSTmap+CVD | 9.06 | 12.17 | **2.23** | 0.25 |

| Proposed Method | **5.57** | **9.50** | 7.69 | **0.75** |
|---|---|---|---|---|

As shown in Tables 3 and 4, because video compression affects the video-based remote physiological signal measurement results, the performance of the methods on the highly compressed dataset COHFACE is significantly lower than that of the uncompressed dataset UBFC-RPPG. Even on unfamiliar datasets with different data distributions, our model still shows stronger generalization ability than other comparison algorithms and provides better results than state-of-the-art methods in terms of MAE and RMSE indicators. In particular, for the highly compressed dataset COHFACE, the overall performance of our method was significantly improved compared to the comparison algorithms. The MAE decreased from 7.8 bpm (CHROM) to 5.57 bpm, and the RMSE dropped below 10 bpm, indicating that our method is more robust in the case of strong equipment noise.

**4.6 Ablation Study**

Table 5 shows our ablation studies on the function model and loss function in different settings on our dataset; the default input video duration was 15 s. Note that “No C_feature extractor” refers to skipped C_feature extractor model, “No Skin Map” refers to canceled skin map generator model, and “Proposed method” includes all the function models and loss functions.

**Evaluation of function module**: To verify the effectiveness of the multilevel feature fusion, we verified the influence of the C_feature extractor and skin map generator in MHFF on the prediction results.

(1) **No C_feature extractor:** To verify the effectiveness of the channel-wise high-level feature map, we skipped the C_feature extractor, which means that the high-level feature map generated by STSC was directly fused with the skin map, and then fed into the softmax classifier to obtain the weight mask $M_w$. As shown in Table 5, after the module was added, the MAE decreased from 3.78 bpm to 2.95 bpm with other settings being the same, and the performance of all other evaluation indicators was improved.

(2) **No skin map:** To verify the effectiveness of the skin map, we canceled the skin map generator and directly sent the channel-wise feature map $f_C$ generated by the C_feature extractor into the softmax classifier to obtain the weight mask $M_w$. As shown in Table 5, the addition of a skin map is important to accurately predict HR. After adding the skin map, the MAE decreased from 7.16 bpm to 2.95 bpm with other settings being the same, and the performance of all other evaluation indicators was considerably improved.

**Evaluation of loss function**: To evaluate the most effective combination of loss functions, we verified the influence of the three loss functions on the prediction results: negative Pearson correlation loss ( $L_r$ ), frequency domain cross-entropy loss ($L_f$), and binary cross-entropy loss ($L_S$). As shown in Table 5, it is sufficient to use only the time domain loss negative Pearson correlation to guide the network to accurately predict the average HR. The joint supervision of the time and frequency domains can enable better HR predictions. The addition of binary entropy loss can guide the skin map generator to predict facial ROI more efficiently and avoid ROI misdetection caused by background noise, which can considerably improve the efficiency of HR prediction.

Table 5. The HR estimation results of the ablation study on our dataset

| Method | MAE↓ | RMSE↓ | $SD_e$↓ | r↑ |
|---|---|---|---|---|
| No C_feature extractor | 1.85 | 3.78 | 4.07 | 0.90 |
| No Skin Map | 5.50 | 7.16 | 9.03 | 0.51 |
| Loss ( $L_r$ ) | 6.88 | 7.18 | 9.95 | 0.44 |
| Loss ( $L_r + L_f$ ) | 5.07 | 7.34 | 8.92 | 0.54 |

| Proposed Method ( $L_r + L_f + L_s$ ) | **1.75** | **3.43** | **2.95** | **0.93** |
|---|---|---|---|---|

## 5. Conclusion

In this study, we proposed a multi-hierarchical perceptual convolutional neural network to obtain color changes in the facial skin regions related to physiological signals from a 15s video. Then, the proposed Signal Predictor (SP) module maps these variations to the rPPG signal and HR. To moderate the deviations of the located facial skin region due to the head motion, a binary skin label with a sparse optical flow pyramid is proposed to achieve high precision motion compensation. The performance of our system is competitive with the state-of-the-art and provides a new solution for reconstructing complex physiological signals from facial videos, which will bring in new technical methods with greater practical value in the fields of telemedicine and emotional computing. However, limited by the complexity and difficulty of physiological signal ground truth acquisition, the effects of light intensity variations were not considered, which will be investigated in our future work.

**Acknowledgements**

The dataset collection has been approved by the local ethics committee, and all the subjects signed an informed consent form. This work was supported by the China Postdoctoral Science Foundation (Program No. 2020M683696XB); the Natural Science Basic Research Plan in Shaanxi Province of China (Program No.2021JQ-455).